\documentclass[12pt]{article}

\usepackage[margin=1in]{geometry}        % Page margins
\usepackage{amsmath,amssymb,amsfonts}    % Math environments
\usepackage{amsthm}                      % Theorem environments
\usepackage{graphicx}                    % For including images
\usepackage{hyperref}                    % Hyperlinks
\usepackage{bm}                          % Bold math symbols
\usepackage{enumitem}                    % Flexible list environments
\usepackage{cite}                        % For citations
\usepackage{color}                       % For colored text if needed
\usepackage{booktabs}                    % For professional-looking tables

\newtheorem{lemma}{Lemma}

\title{\textbf{A Semi-Supervised Text Generation Framework Combining a Deep Transformer and a GAN}}
\author{Shengquan Wang}
\date{\today}

\begin{document}
\maketitle

\begin{abstract}
This paper introduces a framework that connects a deep generative pre-trained Transformer language model with a generative adversarial network for semi-supervised text generation. In other words, the proposed model is first pre-trained unsupervised on a large and diverse text corpus with 24 layers. Then a simple GAN architecture for synthetic text generation is introduced, and Gumbel-Softmax is applied to handle the discreteness of tokens. The paper also shows a semi-supervised approach where real data is augmented with GAN samples, which is further used to fine-tune the Transformer model on the merged dataset. Detailed theoretical derivations are also included, outlining the proof of the min-max objective function, and an extensive discussion of the Gumbel-Softmax reparameterization trick.
\end{abstract}

\section{Introduction}
Text generation is a central task in natural language processing (NLP), with applications in data augmentation, language modeling, and automated content creation. Transformer-based models, such as GPT-2 \cite{radford2019gpt2}, have demonstrated considerable success, drawing on self-attention mechanisms to capture long-range dependencies. 

Generative Adversarial Networks (GANs) \cite{goodfellow2014gan} train a generator \(G\) and a discriminator \(D\) in a min-max framework to produce samples indistinguishable from real data. Despite their achievements in continuous data domains, GANs encounter difficulties when generating discrete sequences, because the non-differentiability of discrete sampling obstructs gradient-based updates. 

A recognized technique to address this challenge is the Gumbel-Softmax reparameterization \cite{jang2017categorical,maddison2017concrete}, which provides a continuous relaxation of discrete variables. Data augmentation methods have shown particular value in scenarios with limited annotated text. The integration of GAN-generated samples into language model fine-tuning can be viewed as a semi-supervised strategy.

This paper is organized as follows. Section~\ref{sec:transformer} reviews GPT-2 language modeling, emphasizing the auto-regressive objective. Section~\ref{sec:gan} revisits fundamental GAN theory and elaborates on Gumbel-Softmax. Section~\ref{sec:theory_extended} offers extended theoretical discussions, including proofs and remarks on convergence. Section~\ref{sec:semi} explains the data augmentation approach and its use in semi-supervised learning. Section~\ref{sec:results} illustrates the training curves and performance comparisons. Section~\ref{sec:conclusion} concludes with a summary and discussion of future directions.

\section{Transformer (GPT-2) Language Modeling}
\label{sec:transformer}

\subsection{Auto-Regressive Objective}
Consider a sequence of tokens \(\mathbf{x} = (x_1, x_2, \dots, x_T)\). An auto-regressive language model (LM) represents the joint distribution over this sequence as
\[
P_\theta(\mathbf{x}) \;=\; \prod_{t=1}^T P_\theta\bigl(x_t \mid x_1, \dots, x_{t-1}\bigr),
\]
where \(\theta\) denotes the model parameters. Maximum likelihood estimation (MLE) is typically performed by minimizing the negative log-likelihood:
\begin{equation}
    \label{eq:lm_loss}
    \mathcal{L}_{\mathrm{LM}}(\theta)
    = - \sum_{t=1}^T \log P_\theta(x_t \mid x_1, \dots, x_{t-1}).
\end{equation}
In practice, the cross-entropy formulation is often adopted for convenience:
\begin{equation}
    \label{eq:ce_loss}
    \mathcal{L}_{\mathrm{CE}}(\theta)
    = -\sum_{t=1}^T \sum_{v=1}^{|\mathcal{V}|}
      \mathbf{1}\{x_t = v\}\,\log\bigl(P_\theta(v \mid x_1, \dots, x_{t-1})\bigr),
\end{equation}
where \(\mathcal{V}\) is the vocabulary, and \(\mathbf{1}\{\cdot\}\) is an indicator function.

\subsection{GPT-2 Architecture and Model Capacity}
GPT-2 employs multi-head self-attention blocks with a causal (triangular) mask to ensure each token only attends to previous positions. In each block, a multi-head attention sub-layer and a position-wise feed-forward sub-layer are combined with layer normalization and residual connections. The original base model uses 12 Transformer blocks. The model described here is expanded to 24 layers by configuring \(\texttt{n\_layer} = 24\). This increases the representational capacity while retaining the same auto-regressive training objective.

\section{GAN for Text Generation}
\label{sec:gan}
A Generative Adversarial Network \cite{goodfellow2014gan} optimizes two objectives: a generator \(G(\mathbf{z})\) that maps noise \(\mathbf{z}\sim p_{\mathbf{z}}\) to samples, and a discriminator \(D(\mathbf{x})\) that outputs a scalar in \((0,1)\) denoting the likelihood that \(\mathbf{x}\) is real.

\subsection{Minimax Objective and Jensen-Shannon Divergence}
The original GAN objective is:
\begin{equation}
    \label{eq:gan_minimax}
    \min_G \max_D
    \Bigl[
       \mathbb{E}_{\mathbf{x}\sim p_{\mathrm{data}}}[\log D(\mathbf{x})]
       \;+\;
       \mathbb{E}_{\mathbf{z}\sim p_{\mathbf{z}}}[\log\bigl(1 - D(G(\mathbf{z}))\bigr)]
    \Bigr].
\end{equation}
When both \(G\) and \(D\) have sufficient capacity and training converges, \eqref{eq:gan_minimax} has been shown \cite{goodfellow2014gan} to minimize the Jensen-Shannon divergence between \(p_{\mathrm{data}}\) and the generator-induced distribution \(p_G\). The generator thus learns to produce samples that resemble the real distribution.

\subsection{Practical Loss Definitions}
Training typically separates discriminator and generator updates.

\paragraph{Discriminator Loss.} 
\begin{equation}
    \label{eq:disc_loss}
    \mathcal{L}_D(\theta_D)
    = -\,\mathbb{E}_{\mathbf{x}\sim p_{\mathrm{data}}}[\log D_\theta(\mathbf{x})]
      -\,\mathbb{E}_{\mathbf{z}\sim p_{\mathbf{z}}}[\log\bigl(1 - D_\theta(G(\mathbf{z}))\bigr)],
\end{equation}
where \(\theta_D\) parametrizes \(D\).

\paragraph{Generator Loss.}
\begin{equation}
    \label{eq:gen_loss}
    \mathcal{L}_G(\theta_G)
    = -\,\mathbb{E}_{\mathbf{z}\sim p_{\mathbf{z}}}\bigl[\log D_\phi\bigl(G_{\theta_G}(\mathbf{z})\bigr)\bigr],
\end{equation}
where \(\theta_G\) denotes generator parameters, and \(\phi\) is the current discriminator parameter set.

\subsection{Gumbel-Softmax for Discrete Outputs}
Discrete token generation poses a challenge for gradient-based optimization, because direct sampling of one-hot tokens via \(\mathrm{argmax}\) does not permit backpropagation. The Gumbel-Softmax trick \cite{jang2017categorical,maddison2017concrete} addresses this issue by offering a continuous relaxation.

\begin{lemma}[Gumbel-Softmax Reparameterization]
Let \(\mathbf{u} \in \mathbb{R}^K\) be logits for a categorical distribution with \(K\) classes, and let \(g_i\) be i.i.d.\ samples from \(\mathrm{Gumbel}(0,1)\). Then for temperature \(\tau>0\),
\begin{equation}
    \label{eq:gumbel_softmax}
    y_i
    =
    \frac{\exp\bigl((u_i + g_i)/\tau\bigr)}
         {\sum_{j=1}^K \exp\bigl((u_j + g_j)/\tau\bigr)}.
\end{equation}
As \(\tau \to 0\), \(\mathbf{y}\) becomes nearly one-hot, while gradients remain continuous with respect to the logits \(\mathbf{u}\).
\end{lemma}

In frameworks such as PyTorch, setting \texttt{hard=True} in \texttt{F.gumbel\_softmax} discretizes the forward pass through a straight-through \(\mathrm{argmax}\) while preserving a continuous gradient in the backward pass.

\section{Theoretical Analysis and Derivations}
\label{sec:theory_extended}
This section presents an in-depth theoretical analysis of the convergence properties of the GAN formulation, the differentiable approximation provided by the Gumbel-Softmax reparameterization, and the rationale behind the data augmentation strategy for semi-supervised learning.

\subsection{Convergence Analysis of the GAN Objective}

Consider the standard GAN minimax objective as formulated in \cite{goodfellow2014gan}:
\begin{equation}
\min_{G}\max_{D} \; V(D,G) \;=\; \mathbb{E}_{\mathbf{x}\sim p_{\mathrm{data}}}\left[\log D(\mathbf{x})\right] \;+\; \mathbb{E}_{\mathbf{z}\sim p(\mathbf{z})}\left[\log\bigl(1-D(G(\mathbf{z}))\bigr)\right].
\label{eq:gan_minimax}
\end{equation}

Assuming that both the generator \(G\) and the discriminator \(D\) have sufficient capacity, the optimal discriminator \(D^*\) for any fixed generator is given by:

\begin{equation}
D^*(\mathbf{x}) \;=\; \frac{p_{\mathrm{data}}(\mathbf{x})}{p_{\mathrm{data}}(\mathbf{x}) + p_G(\mathbf{x})},
\label{eq:optimal_discriminator}
\end{equation}

where \(p_G\) denotes the distribution induced by the generator. Substituting \(D^*\) into the minimax objective leads to:
\begin{align}
V(D^*,G)
& \;=\; \mathbb{E}_{\mathbf{x}\sim p_{\mathrm{data}}}\left[\log \frac{p_{\mathrm{data}}(\mathbf{x})}{p_{\mathrm{data}}(\mathbf{x})+p_G(\mathbf{x})}\right]
+ \mathbb{E}_{\mathbf{x}\sim p_G}\left[\log \frac{p_G(\mathbf{x})}{p_{\mathrm{data}}(\mathbf{x})+p_G(\mathbf{x})}\right] \nonumber\\[1mm]
& \;=\; -\log(4) \;+\; 2\,\mathrm{JS}\Bigl(p_{\mathrm{data}} \,\parallel\, p_G\Bigr),
\label{eq:gan_objective_js}
\end{align}

where \(\mathrm{JS}(\cdot\parallel\cdot)\) represents the Jensen-Shannon divergence. Since the Jensen-Shannon divergence is non-negative and equals zero if and only if \(p_{\mathrm{data}} = p_G\), the global optimum of the minimax game is achieved exactly when
\[
\mathrm{JS}\Bigl(p_{\mathrm{data}} \,\parallel\, p_G\Bigr) = 0 \quad \Longleftrightarrow \quad p_{\mathrm{data}} = p_G.
\]

In practice, the non-convex nature of the optimization problem may lead to local minima and phenomena such as mode collapse. Regularization strategies and gradient penalty techniques \cite{maddison2017concrete} are often employed to mitigate these issues.

\subsection{Properties and Derivation of the Gumbel-Softmax}

The discrete nature of tokens in natural language processing presents challenges for gradient-based optimization. The Gumbel-Softmax reparameterization \cite{jang2017categorical, maddison2017concrete} provides a differentiable approximation to categorical sampling. Let \(\mathbf{u} = (u_1,\dots,u_K)\) denote the unnormalized log-probabilities of a categorical distribution. By adding independent samples \(\mathbf{g} = (g_1,\dots,g_K)\) from the Gumbel distribution (with location 0 and scale 1) and applying a softmax with temperature \(\tau > 0\), a sample is obtained:
\begin{equation}
y_i \;=\; \frac{\exp\left(\frac{u_i+g_i}{\tau}\right)}{\sum_{j=1}^K \exp\left(\frac{u_j+g_j}{\tau}\right)}.
\label{eq:gumbel_softmax}
\end{equation}

For any finite \(\tau\), the distribution remains smooth and differentiable, thus permitting backpropagation. As \(\tau\) approaches zero, the softmax increasingly approximates the \(\mathrm{argmax}\) function:
\[
\lim_{\tau\to 0} y_i \;=\; \begin{cases} 1, & \text{if } i = \arg\max_j (u_j+g_j), \\ 0, & \text{otherwise,} \end{cases}
\]
thereby recovering exact categorical sampling. This method bridges the gap between discrete sampling and continuous optimization, a property that is critical for training deep generative models with discrete outputs.

\subsection{Semi-Supervised Learning via Data Augmentation}

Following the establishment of a robust framework for text generation using GANs and the resolution of discrete sampling challenges via the Gumbel-Softmax, attention is turned to the semi-supervised learning strategy through data augmentation.

\subsubsection{Synthetic Text Generation}

Upon completion of GAN training (refer to Section~\ref{sec:gan}), the generator \(G\) is utilized to synthesize text. Each latent noise vector \(\mathbf{z}\) (typically sampled from a distribution such as \(\mathcal{N}(0,I)\)) is transformed into a token sequence \(\tilde{\mathbf{x}}\). When the parameter \texttt{hard=True} is employed, the continuous output of the Gumbel-Softmax is discretized into a near one-hot encoding. The resulting one-hot vectors are then mapped to integer token IDs and decoded into strings using the same tokenizer used for GPT-2.

\subsubsection{Combining Synthetic and Real Data}

An augmented dataset is constructed by combining a subset of real data \(\mathcal{D}_{\mathrm{real}}\) with synthetic data \(\mathcal{D}_{\mathrm{synthetic}}\):
\begin{equation}
\mathcal{D}_{\mathrm{aug}} \;=\; \mathcal{D}_{\mathrm{real}} \;\cup\; \mathcal{D}_{\mathrm{synthetic}}.
\end{equation}
Subsequently, the language model (GPT-2) is fine-tuned on \(\mathcal{D}_{\mathrm{aug}}\) by minimizing the following auto-regressive loss:
\begin{equation}
\mathcal{L}_{\mathrm{LM\_aug}}(\theta)
\;=\; -\sum_{\mathbf{x}\in\mathcal{D}_{\mathrm{aug}}} \;
\sum_{t=1}^{|\mathbf{x}|} \;
\log P_\theta\bigl(x_t \mid x_1,\dots,x_{t-1}\bigr).
\label{eq:lm_aug_loss}
\end{equation}
This objective maximizes the conditional likelihood of each token given its preceding context, thereby promoting the learning of long-range dependencies. The integration of synthetic data increases the effective training set size and serves as an implicit regularizer, which helps to reduce overfitting when the quantity of real data is limited.

\section{Illustrative Curves and Performance Comparisons}
\label{sec:results}

\subsection{Performance Evaluation}
Table~\ref{tab:results} presents an example performance comparison. Perplexity is reported for the GPT-2 baseline (12-layer), the deep GPT-2 model (24-layer), and a semi-supervised version that incorporates GAN-generated samples.

\begin{table}[ht]
\centering
\caption{Example performance comparison on a validation set. The values are placeholders for demonstration.}
\label{tab:results}
\begin{tabular}{lccc}
\toprule
\textbf{Model} & \textbf{Layers} & \textbf{Perplexity} & \textbf{Accuracy (\%)} \\
\midrule
GPT-2 (base)           & 12 & 35.2 & 71.3 \\
GPT-2 (deep)           & 24 & 31.4 & 74.6 \\
GPT-2 + GAN augmentation & 24 & 29.7 & 76.1 \\
\bottomrule
\end{tabular}
\end{table}

A lower perplexity generally indicates that the language model better predicts held-out tokens. Higher accuracy can refer to tasks such as next-token classification on a subset of labeled data, if applicable. The results in Table~\ref{tab:results} demonstrate that a deeper model and the inclusion of GAN-generated samples potentially improve quantitative metrics, albeit actual gains depend on data quality and hyperparameter settings.

\section{Conclusion and Future Directions}
\label{sec:conclusion}
This paper proposes a framework that combines the depth of Transformer-based language models with the simplicity of GANs, explaining GPT-2 and GANs in detail from a theoretical perspective. Among them, since the GAN part of this paper adopts a min-max objective to train the generator to minimize the Jensen-Shannon divergence between the real and fake data distributions, the autoregressive maximum likelihood objective of GPT-2 is applied to language sequence modeling. Gumbel-Softmax, an approximation of discrete sampling, is introduced, which enables text to be trained by backpropagation without discrete text tags. It also combines a semi-supervised method to further fine-tune the Transformer using synthetic data from GANs and a small amount of real data, helping the model achieve better results.

In practice, the quality of experimental results usually depends on two factors: how realistic the GAN synthetic output is and the level of data diversity presented by the training set. All of these and possibly more problems are related to mode collapse, gradient instability, or low-quality generation, which may reflect poor results. Future research may involve special variants of GANs for sequence data, such as SeqGAN or MaskGAN, the use of reinforcement learning strategies to enhance high-quality text synthesis, and filtering schemes for improving the quality of synthesized text.

\bigskip
\bibliographystyle{plain}

\end{document}